\documentclass[conference]{IEEEtran}
\IEEEoverridecommandlockouts
\usepackage{cite}
\usepackage{amsmath,amssymb,amsfonts}
\usepackage{algorithmic}
\usepackage{graphicx}
\usepackage{textcomp}
\usepackage{xcolor}
\usepackage{hyperref}
\usepackage{graphicx}
\usepackage{amsthm}
\usepackage{booktabs}
\usepackage{multirow}
\usepackage{makecell}
\usepackage{algorithm}
\usepackage[switch]{lineno}
\usepackage{subfigure}
\usepackage{pifont}
\usepackage{soul}

\usepackage{color, colortbl}

\usepackage{mathtools}
\DeclareMathOperator*{\concat}{%
    \mathchoice%
        {\Big\Vert}%
        {\big\Vert}%
        {\Vert}%
        {\Vert}%
}

\def\BibTeX{{\rm B\kern-.05em{\sc i\kern-.025em b}\kern-.08em
    T\kern-.1667em\lower.7ex\hbox{E}\kern-.125emX}}
\begin{document}

\title{Trustworthy Enhanced Multi-view Multi-modal Alzheimer's Disease Prediction with Brain-wide Imaging Transcriptomics Data}

\author{\IEEEauthorblockN{Shan Cong$^{1,2}$,
Zhoujie Fan$^{1,2}$,
Hongwei Liu$^1$,
Yinghan Zhang$^3$,
Xin Wang$^3$,
Haoran Luo$^1$\IEEEauthorrefmark{1}, and
Xiaohui Yao$^{1,2}$\IEEEauthorrefmark{1}}
\IEEEauthorblockA{$^1$\textit{College of Intelligent Systems Science and Engineering, Harbin Engineering University, Harbin, China}}
\IEEEauthorblockA{$^2$\textit{Qingdao Innovation and Development Center, Harbin Engineering University, Qingdao, China}}
\IEEEauthorblockA{$^3$\textit{Department of Surgery, Chinese University of Hong Kong, Hong Kong, China}}
\thanks{Corresponding authors: Xiaohui Yao (email: xiaohui.yao@hrbeu.edu.cn); Haoran Luo (email: luohaoran@hrbeu.edu.cn).}}

\maketitle

\begin{abstract}
Brain transcriptomics provides insights into the molecular mechanisms by which the brain coordinates its functions and processes. 
However, existing multimodal methods for predicting Alzheimer's disease (AD) primarily rely on imaging and sometimes genetic data, often neglecting the transcriptomic basis of brain. 
Furthermore, while striving to integrate complementary information between modalities, most studies overlook the informativeness disparities between modalities. 
Here, we propose TMM, a trusted multiview multimodal graph attention framework for AD diagnosis, using extensive brain-wide transcriptomics and imaging data. 
First, we construct view-specific brain regional co-function networks (RRIs) from transcriptomics and multimodal radiomics data to incorporate interaction information from both biomolecular and imaging perspectives. 
Next, we apply graph attention (GAT) processing to each RRI network to produce graph embeddings and employ cross-modal attention to fuse transcriptomics-derived embedding with each imaging-derived embedding. Finally, a novel true-false-harmonized class probability (TFCP) strategy is designed to assess and adaptively adjust the prediction confidence of each modality for AD diagnosis. 
We evaluate TMM using the AHBA database with brain-wide transcriptomics data and the ADNI database with three imaging modalities (AV45-PET, FDG-PET, and VBM-MRI). The results demonstrate the superiority of our method in identifying AD, EMCI, and LMCI compared to state-of-the-arts.
Code and data are available at \href{https://github.com/Yaolab-fantastic/TMM}{https://github.com/Yaolab-fantastic/TMM}.
\end{abstract}



\section{Introduction} 
Alzheimer's disease (AD) is a progressive neurodegenerative disorder that predominantly affects the elderly, characterized by the gradual deterioration of cognitive functions, such as memory, reasoning, and decision-making~\cite{jucker2023alzheimer}. 
Studies have demonstrated that the pathological changes associated with AD begin to manifest years or even decades before the earliest clinical symptoms are observed~\cite{bateman2012clinical,yao2019bioinfo}. 

Recent advances in acquiring biomedical data, coupled with the rapid advancements in machine learning and deep learning techniques, have significantly enhanced the development of computer-aided diagnosis of AD. The multifactorial nature of AD highlights the need for multi-modal learning, where numerous algorithms have been developed to capture and integrate complementary information from various modalities. Existing multimodal methods for AD prediction primarily focus on using brain imaging data~\cite{qiu2022multimodal,CHEN2023102698,QiuTMI10498133,zhou2020multi,zhang2020multi}. 
For example, Qiu et al.~\cite{QiuTMI10498133} developed a local-aware convolution to capture intra-modality associations and employed a self-adaptive Transformer (SAT) to learn inter-modality global relationships. Zhang et al.~\cite{zhang2020multi} proposed an attention-based fusion framework to selectively extract features from MRI and PET branches.

Although imaging modalities effectively quantify the structural and functional alterations in the brain, they do not account for the molecular mechanisms underlying these changes.
A few studies have started to incorporate genotyping data to enhance model performance~\cite{ShenLimiccai2023,WANG2023102883,ko2022deep,bi2022novel}. For example, Zhou et al.~\cite{ShenLimiccai2023} proposed an attentive deep canonical correlation analysis to integrate imaging modalities with genetic SNP data for diagnostic purposes. While genotypic data can characterize individuals, this upstream information does not directly explain cellular function and pathological states. \textit{Compared to genetics, downstream transcriptomics provides more direct insights by reflecting the actual activity and expression of genes within specific tissue contexts. This motivates us to incorporate brain-specific transcriptomics with imaging data into AD prediction.}

Besides improving representation learning, there has been a growing emphasis on developing various fusion strategies to capture the complementary information present across modalities~\cite{zhu2022deep,song2021graph,yao2024mocat,bi2023community,liang2024gremi}. For example, Song et al.~\cite{song2021graph} developed functional and structural graphs using fMRI and DTI images, and combined the multimodal information into edges through an efficient calibration mechanism. Bi et al.~\cite{bi2023community} developed a community graph convolutional neural network to model brain region-gene interactions, using an affinity aggregation model to enhance interpretability and classification performance in AD diagnosis. Although effective, existing multimodal algorithms often lack a dynamical perception of the informativeness of each modality for different samples, which could otherwise enhance the trustworthiness, stability, and explainability of these methods. 
\textit{Evidence has shown that the informativeness of a modality typically varies across different samples~\cite{rideaux2021multisensory}, which motivates efforts to model modality informativeness to enhance fusion effectiveness~\cite{Han2022MultimodalDD,luo2024teminet,Zheng2023multilevel}.}

\begin{figure*}[th]
\centering
\includegraphics[width=0.8\textwidth]{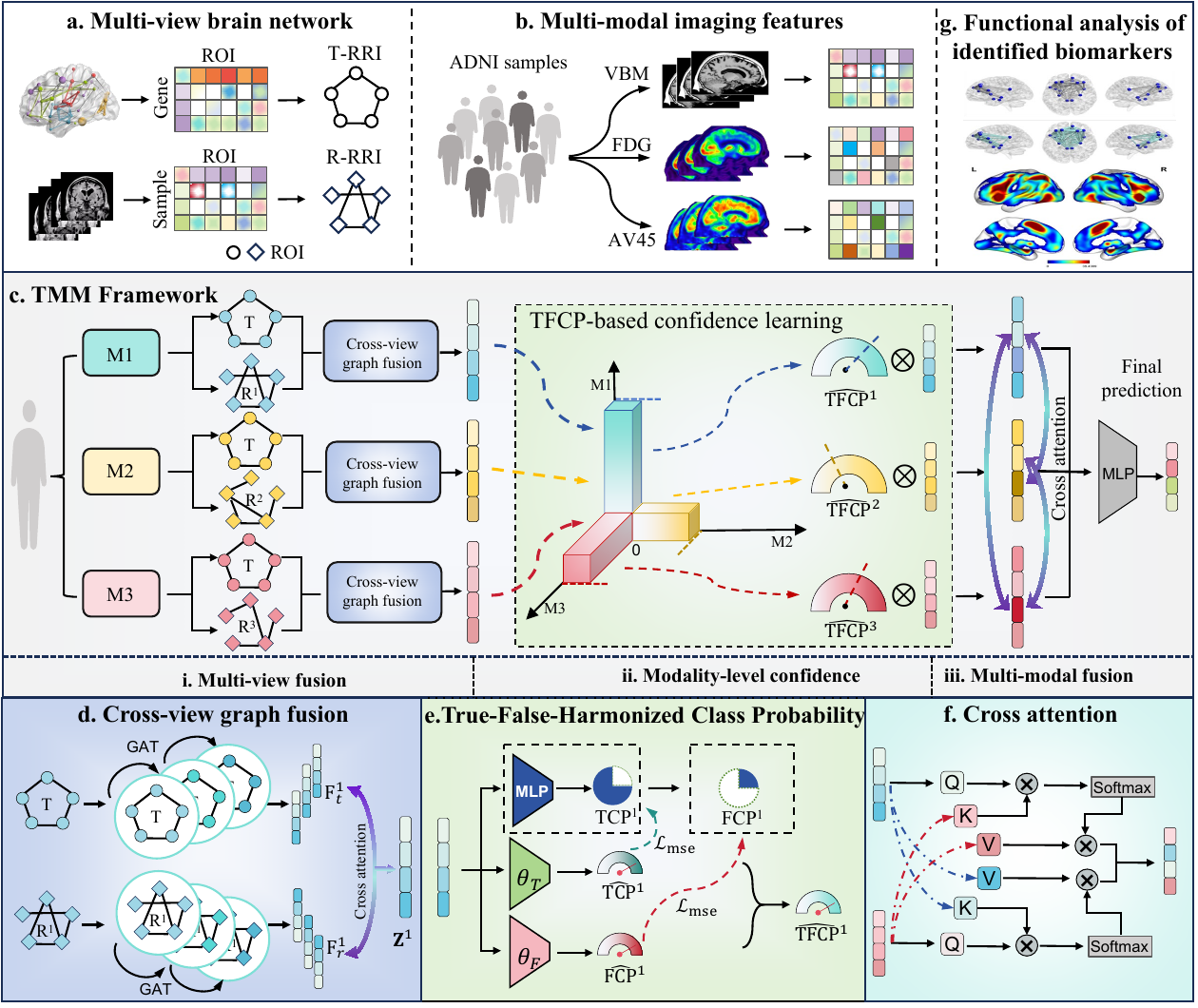}
\caption{Framework overview. (a) Transcriptomics-specific RRI network (T-RRI) is constructed from the brain-wide gene expression data, and radiomics-specific RRI network is derived from each imaging modality. (b) Three imaging modalities, including VBM, FDG, and AV45, are employed in our analysis. (c) The TMM architecture: (i) For each sample, modality-specific T-RRI and R-RRI networks are constructed by mapping ROI measurements to each node. Multi-level GAT is applied to capture interactions within ROIs for each view, generating view-specific representations; followed by cross-view attention to create multi-view embeddings. (ii) TFCP-based confidence networks are designed to estimate the prediction confidence of each modality. (iii) Cross-modal attention mechanisms fuse multiple modalities to deliver the final predictive output. (d-f) illustrates the details of cross-view graph fusion, the TFCP mechanism, and the cross-attention mechanisms, respectively. (g) The identified ROI biomarkers are annotated for their functional connection and co-activation.}\label{fig:framework}
\end{figure*}

To address the aforementioned issues, we propose a trustworthy enhanced multi-view multi-modal graph network (TMM) that integrates brain-wide transcriptomics knowledge and multi-modal radiomics data for predicting AD in an adaptive and trustworthy manner. In this context, multi-view refers to the incorporation of both transcriptomics and radiomics data, while multi-modal pertains to including different imaging modalities. 
As shown in Figure~\ref{fig:framework}, we begin by constructing view-specific region of interest (ROI) co-function networks (RRIs) from brain-wide gene expression data and brain imaging data, to capture the molecular, structural, and functional relationships underlying different brain regions. 
Graph attention networks (GAT) are then applied to generate sample-wise embeddings, followed by cross-attention to fuse multi-view representations. 
In the fusion stage, we propose a novel true-false-harmonized class probability (TFCP) criterion to measure modality confidence, facilitating the adaptive perception of and response to variations in modality informativeness.
Our main contributions can be summarized as follows:
\begin{itemize}
    \item We propose a multi-view, multi-modal learning framework for AD prediction that integrates brain-specific transcriptomics and imaging data. This approach innovatively models the co-functional relationships of ROIs by combining insights from upstream molecular activities and downstream anatomical and functional characteristics.
    \item We propose a novel modality confidence learning strategy that estimates the harmonized true and false class probability to dynamically assess the modality informativeness. This design enables a more robust and trustworthy multimodal prediction. 
    \item Our model significantly outperforms leading methods in multiple prediction tasks. A series of ablation studies robustly validate the effectiveness of the proposed TMM framework. Furthermore, we identify important ROIs and demonstrate their functional roles in brain cognition.
\end{itemize}

\section{Methods}

Let $\mathbf{T}= \{\mathbf{t}_1, \dots, \mathbf{t}_d\} \in \mathbb{R}^{n_g\times d}$ represent the brain-wide transcriptomics data, with $n_g$ denoting the number of genes and $d$ denoting the number of ROIs.
Let $m \in [1,\dots, M]$ denotes the $m$-th imaging modality, $\mathbf{X}^{m} = \{\mathbf{x}^m_1, \dots, \mathbf{x}^m_d\} \in  \mathbb{R}^{n\times d}$ represents the $m$-th imaging feature matrix, and $\mathbf{y}=[y_1,\dots,y_n]\in \mathbb{R}^{n}$ denotes the label vector, where $n$ is the number of samples and $d$ is the number of ROIs. 

We now detail the proposed TMM framework. Initially, one transcriptomics-specific network (T-RRI) and $M$ radiomics-specific RRI networks (R-RRIs) are constructed from brain-wide gene expression data and each imaging modality dataset, respectively. For each imaging modality, ROI-level imaging measurements are assigned as node features to both the T-RRI and R-RRI networks, generating graph representations from transcriptomic and imaging views. 
GATs are then applied to each RRI network to generate view-specific embeddings. These embeddings are integrated using cross-modal attention mechanisms to form multi-view representations. Subsequently, a confidence learning network is specifically designed to enhance the reliability of the embeddings, which are further integrated using cross-modal attention for final prediction.
The following sections provide further details on the construction of RRI networks, multi-view representation and fusion, trustworthy evaluation, and multi-modal integration. 
The overall architecture and important modules are illustrated in Figure~\ref{fig:framework}.

\subsection{Sample-wise Co-functional RRI Network Construction}\label{sec:graph_construction}
As shown in Figure~\ref{fig:framework}(a), two types of RRI networks (T-RRI and R-RRI) are constructed from transcriptomics and imaging data, respectively, to model the co-functionality of ROIs and facilitate graph-based network implementation. 
Given the brain-wide gene expression matrix $\mathbf{T} \in \mathbb{R}^{n_g\times d}$, we derive the edge matrix $\mathbf{E}^t =\{e_{ij}^t\} \in \mathbb{R}^{d\times d}$ by thresholding the adjacency matrix, which measures the correlation coefficient for each pair of ROIs across all expressed genes: 
\begin{equation}
\small
\begin{aligned}
\label{eq:adj}
    e_{ij}^t=\left\{\begin{array}{cl}
     1 & \text { if } r(\mathbf{t}_i, \mathbf{t}_j) \geq \lambda_t \\
     0 & \text { otherwise}
    \end{array}\right.,
\end{aligned}
\end{equation}%
where $r(\mathbf{t}_i, \mathbf{t}_j)$ computes the pearson correlation coefficient (PCC) of ROIs $\mathbf{t}_i$ and $\mathbf{t}_j$ expressions, and $\lambda_t$ is a hyperparameter for thresholding.

We apply the same strategy to each imaging modality, obtaining $M$ imaging-specific edge matrix $\mathbf{E}^m =\{e_{ij}^m\}\in\mathbb{R}^{d\times d}$ ($m \in [1,\dots, M]$), with each element $e^m_{i,j}$ representing the thresholded PCC (by $\lambda_r$) between imaging measures of each ROI pair across all $n$ samples. Across experiments, we set the thresholds $\lambda_t=0.2$ for transcriptomics-based networks and $\lambda_r=0.1$ for radiomics-based networks, respectively.

Afterward, for each imaging modality, we devise one T-RRI network and one R-RRI network for each subject by assigning ROI-level imaging measurements to their corresponding nodes as features. 
Consequently, each subject is associated with $M$ T-RRI graphs and $M$ R-RRI graphs. Notably, the $M$ T-RRI graphs share the same edge matrix but differ in node features, while the $M$ R-RRI graphs each have unique edge matrices and distinct node features.

\subsection{Multi-view Fusion for Modal-Specific Representation}\label{sec:graph_fusion}
\textbf{View-specific representation.} Given their ability to effectively aggregate neighboring relationships within graph data, we utilize GATs for multi-view sample representation learning. Specifically, GATs are applied to each T-RRI and R-RRI graph, enabling the sample representations from both transcriptomic and different imaging views. Illustrated with the $m$-th modality, we take $\mathcal{G}^{m}_{t,0}=(\mathbf{X}^{m},\mathbf{E}^t)$ and $\mathcal{G}^{m}_{r,0}=(\mathbf{X}^{m},\mathbf{E}^m)$ as input T-RRI and R-RRI graphs and stack graph attention layers with multi-head attention to build the GAT for each graph (as shown in Figure~\ref{fig:framework}(d)). Each layer is defined as: 
\begin{equation}
\small
\label{eq:gat}
    \boldsymbol{h}_{u}^{\prime}
    =\|_{k=1}^K
    \sigma \left(
    \sum\limits_{v\in {\mathcal{N}_u}} \alpha_{uv}^k \mathbf{W}^k \boldsymbol{h}_v
    \right),
\end{equation}
where $\parallel$ is concatenation of $K$ heads, $\boldsymbol{h}_v$ is the input features of node $v$, $\alpha_{uv}^k$ is the $k$-th normalized attention coefficients, $\mathbf{W}^k$ is the weight matrix of head $k$, $\mathcal{N}_u$ is the first-order neighbors of node $u$, and $\sigma(\cdot)$ is a nonlinear activation function.

We further adopted a multi-level strategy to enhance information aggregation between brain ROIs. 
Specifically, higher-level graphs $\mathcal{G}^{m}_{t,1}$ and $\mathcal{G}^{m}_{r,1}$ are generated by applying GAT to initial graphs $\mathcal{G}^{m}_{t,0}$ and $\mathcal{G}^{m}_{r,0}$. Then, we derived the next level graphs, $\mathcal{G}^{m}_{t,2}$ and $\mathcal{G}^{m}_{r,2}$ from $\mathcal{G}^{m}_{t,1}$ and $\mathcal{G}^{m}_{r,1}$ respectively. 
For each modality, graph embeddings from three levels are concatenated to produce view-specific representations $\mathbf{F}^{m}_{t}$ and $\mathbf{F}^{m}_{r}$.

\textbf{Multi-view fusion.} Within each modality, we introduce cross-view attention mechanisms to integrate representations from the T-RRI and R-RRI networks (as shown in Figure~\ref{fig:framework}(d)), enabling them to complement and enhance each other:
\begin{equation}
\small
\begin{aligned}\label{eq:mvfusion}
  \mathbf{Z}^{m}_{t} &=\text{Att}\Big(\mathbf{F}^m_{t}\mathbf{W}^Q_t, \mathbf{F}^m_{r}\mathbf{W}^K_{r}, \mathbf{F}^m_{r}\mathbf{W}^V_{r}\Big),\\
  \mathbf{Z}^{m}_{r} &= \text{Att}\Big(\mathbf{F}^m_{r}\mathbf{W}^Q_r, \mathbf{F}^m_{t}\mathbf{W}^K_{t}, \mathbf{F}^m_{t}\mathbf{W}^V_{t}\Big),\\
  \mathbf{Z}^{m}&=\mathbf{Z}^{m}_{t}\concat\mathbf{Z}^{m}_{r}.
\end{aligned}
\end{equation}

In addition to fusing transcriptomics and radiomics information, we also train a modal-specific classifier to incorporate within-modal information into prediction.
In particular, for the $m$-th modality, a GAT classifier is trained as:
\begin{equation}
\small
\label{eq:gat_loss}
    \mathcal{L}_{\text{GAT}}^{m}=
    \sum\limits_{i=1}^{n}\mathcal{L}_{\text{CE}}(\hat{y}_i^{m},y_i),
\end{equation}
where $\mathcal{L}_{\text{CE}}$ is the cross-entropy loss, $y_i$ is the true label, and $\hat{y_i^{m}}$ is the prediction from $\mathbf{Z}^m$.

\subsection{Modality-level Confidence: True-False-Harmonized Class Probability}\label{sec:tfcp}
In multimodal learning, data heterogeneity is a common challenge, with the discriminative capability of each modality varying across different samples. Therefore, it is crucial to develop algorithms that can adaptively perceive and respond to these variations in informativeness. 
By transforming the assessment of modal informativeness into an evaluation of the confidence levels associated with modal classification performance, we can effectively estimate the modality informativeness through its prediction confidence. Here, we propose a novel TFCP criterion to approximate the prediction confidence of each modality.

The confidence criterion measures how confident the model is in its predictions by correlating high prediction certainty with greater values and vice versa. For a classifier $f: \mathbf{x}_i \rightarrow y_i$, the traditional confidence criterion utilizes the maximum class probability (MCP): $\text{MCP}(\mathbf{x}_i)=P(\widehat{y_i}|\mathbf{x}_i)$, where $\widehat{y_i}$ is the class with the largest softmax probability. However, it can be observed that this method assigns high confidence (i.e., high softmax probability) to both correct and incorrect predictions, resulting in overconfidence for failure predictions. True class probability (TCP) criterion~\cite{corbiere2019addressing} has been proposed to calibrate incorrect predictions. Instead of assigning the maximum class probability, TCP uses the softmax probability of the true class $y_i^*$ as prediction confidence: $\text{TCP}(\mathbf{x}_i)=P(y_i^*|\mathbf{x}_i)$, where $y_i^*$ is the true class (i.e., the actual label). The TCP strategy exclusively models the certainty of the true class while neglecting the uncertainty of the untrue classes, which could result in a biased and unstable confidence approximation. 

To this end, we propose considering both true and false class probabilities to aggregate evidence from these two perspectives. Specifically, we design a harmonized criterion
for approximating the prediction confidence:
\begin{equation}
\small
\begin{aligned}
\label{eq:tfcp}
    \text{TFCP}(\mathbf{x}_i) 
    &= \frac{2}{1/{\text{TCP}(\mathbf{x}_i)} + 1/({1-\text{FCP}(\mathbf{x}_i)})}\\
    &= \frac{2}{1/{P(y_i^*|\mathbf{x}_i)} + 1/{(1-P(\overline {y_i^*}|\mathbf{x}_i))}},
\end{aligned}
\end{equation}%
where $\text{FCP}(\mathbf{x}_i)=P(\overline {y_i^*}|\mathbf{x}_i)$ is the softmax probability of false (or untrue) class. 

It should be noted that neither the TCP nor the FCP can be directly applied to test samples because they both require label information. To address this issue, we introduce two confidence networks to approximate the TCP and FCP, respectively. As shown in Figure~\ref{fig:framework}(e), a TCP confidence network with parameters $\theta_T$ and an FCP confidence network with parameters $\theta_F$ are constructed based on the sample representations to generate certainty and uncertainty scores. These networks are trained to minimize the discrepancy between predicted and actual scores:
\begin{equation}
\label{eq:loss_tfcp}
\mathcal{L}_{\mathrm{Conf}}^{m} \!=\! \left\Vert  \text{TFCP}(\mathbf{Z}^{m})
- \widehat{\text{TFCP}}(\mathbf{Z}^{m}, \theta^{m}_{T}, \theta^{m}_{F}) \right\Vert^2 \!+\! \mathcal{L}_{\text{Cls}}^{m}.
\end{equation}
Here, both the TCP and FCP networks are built upon a classification network trained using the cross-entropy loss $\mathcal{L}_{\text{Cls}}^{m}$.

\subsection{Multi-modal Fusion: Cross-modal 
Attention}\label{sec:mmfusion}
From the previous sections, multi-view sample representations for each modality have been derived, and modality informativeness has also been assessed, denoted as $\mathbf{Z}^{m}$ and $\widehat{\text{TFCP}}^{m}$, respectively. By incorporating the modality confidence into the corresponding representations, we can obtain trustworthy representation for each modality:
\begin{equation}
\begin{aligned}
\label{eq:trustrep}
\mathbf{H}^{m} = \widehat{\text{TFCP}}^{m} \cdot \mathbf{Z}^{m}
\end{aligned}
\end{equation}

Now, we employ cross-modal attention mechanisms to enhance each modality by leveraging insights from others, subsequently concatenating the enriched modal representations for final prediction:
\begin{equation}
\small
\begin{aligned}
\label{eq:mmatt}
\mathbf{U}^{m} &=  \|_{j=1,j\neq m}^M \text{Att}\Big( 
\mathbf{H}^{m}\mathbf{W}^{Q}_{m}, \mathbf{H}^{j}\mathbf{W}^{K}_{j}, \mathbf{H}^{j}\mathbf{W}^{V}_{j}
\Big),\\
\mathbf{U}&=\|_{m=1}^M \mathbf{U}^{m}.
\end{aligned}
\end{equation}

\subsection{Objective optimization.} The overall loss is composed of the modality-specific classification loss (Eq.~\ref{eq:gat_loss}), the TFCP loss (Eq.~\ref{eq:loss_tfcp}), and the cross-entropy loss of the final classification:
\begin{equation}\label{eq:loss_total}
\small
   \mathcal{L}= \eta_{1}\sum\limits_{m=1}^{M}\mathcal{L}_{\text{GAT}}^{m}+ \eta_{2}\sum\limits_{m=1}^{M}\mathcal{L}_{\text{Conf}}^{m}+
   {\mathcal{L}_{\text{Final}}},
\end{equation}
where $\eta_{1}$ and $\eta_{2}$ denote hyperparameters for adjusting different losses. We set $\eta_{1}=1$  and $\eta_{2}=1$  across our experiments.

\section{Experiments and Results}

\subsection{Dataset}
\textbf{AHBA dataset.} Brain-wide transcriptomics data are sourced from the Allen human brain atlas (AHBA, \url{human.brain-map.org}), including over 58k probes sampled across 3,702 brain locations from six donors. 
We use the abagen toolbox~\cite{abagen} for data preprocessing and employ the AAL atlas to map brain locations to ROIs, resulting in a number of 15,633 genes expressed across 116 ROIs. 
To ensure that the prior knowledge introduced closely aligns with the underlying mechanisms of AD, we further filter genes based on a large-scale AD meta-GWAS from the IGAP~\cite{kunkle2019genetic}. Specifically, we use the MAGMA~\cite{de2015magma} to derive gene-level $p$-values from SNP-level GWAS and keep nominally significant ones (i.e., $p<$0.05). This process yields a total of 1,216 genes across 116 ROIs for constructing the T-RRI edge matrix.

\textbf{ADNI dataset.} We conduct experimental validation using subjects from the Alzheimer's Disease Neuroimaging Initiative (ADNI, \url{https://adni.loni.usc.edu/}).
Multimodal imaging data--including AV45-PET, FDG-PET, and VBM-MRI, along with corresponding diagnosis labels (NC, EMCI, LMCI, and AD) for each visit, are collected from 851 ADNI participants. Each scan is processed according to established protocols~\cite{yao2019bioinfo}, and measurements for 116 ROIs are derived from each modality using the AAL atlas. 
Details are listed in Table~\ref{tab:dataset}.
\def\arraystretch{0.7}
\begin{table}[!tbh]
\centering
\caption{Demographic characteristics of the ADNI subjects.\label{tab:dataset}}
\begin{tabular}{lcccc}
\toprule
Subjects    & NC    & EMCI  & LMCI  & AD \\\midrule
Number      & 221    & 275    & 190    & 165 \\
Sex(M/F)    & 114/107    & 155/120    & 111/79    & 99/66 \\
Age         & 76.34±6.58    & 71.50±7.12    & 74.08±8.55    & 75.35±7.88 \\
Education   & 16.40±2.66    & 16.09±2.61    & 16.36±2.79    & 15.87±2.72 \\
\bottomrule
\addlinespace
\end{tabular}
\end{table}


\begin{table*}[!thb]
\centering
\caption{The comparison results on four ADNI classification tasks.}\label{tab:sota}
\begin{tabular}
{@{}>{\centering\arraybackslash}p{35pt}|>{\centering\arraybackslash}p{25pt}>{\centering\arraybackslash}p{25pt}>{\centering\arraybackslash}p{25pt}|>{\centering\arraybackslash}p{25pt}>{\centering\arraybackslash}p{25pt}>{\centering\arraybackslash}p{25pt}|>{\centering\arraybackslash}p{25pt}>{\centering\arraybackslash}p{25pt}>{\centering\arraybackslash}p{25pt}|>{\centering\arraybackslash}p{25pt}>{\centering\arraybackslash}p{25pt}>{\centering\arraybackslash}p{25pt}}
\toprule
\multirow{3}{*}{Method} & \multicolumn{3}{c|}{NC vs. AD} & \multicolumn{3}{c|}{NC vs. LMCI}& \multicolumn{3}{c|}{NC vs. EMCI} & \multicolumn{3}{c}{EMCI vs. LMCI} \\ 
\cmidrule{2-4}  \cmidrule{5-7} \cmidrule{8-10}  \cmidrule{11-13}
& ACC & F1 & AUC & ACC & F1 & AUC & ACC & F1 & AUC & ACC & F1 & AUC  \\ \midrule 
SVM                  & 90.2±2.3           & 89.8±1.2           & 94.1±4.2           & 65.4±2.3           & 63.5±2.5           & 72.1±3.3          							& 67.7±1.2           & 78.5±1.4           & 64.9±1.7          & 62.4±3.1           & 69.3±3.0          & 65.5±3.0          								\\
Lasso                & 91.2±3.0           & 89.8±3.6           & 96.6±1.4           & 67.7±4.6           & 65.6±5.5           & 76.7±3.1          							& 66.5±3.4           & 76.2±2.6           & 66.7±3.2          & 65.4±3.2           & 70.7±3.8          & 68.0±2.0          								\\
XGBoot               & 91.2±1.4           & 89.7±1.6           & \underline {96.8±0.8}     & 70.1±1.2           & 67.3±1.9           & 76.0±1.5   							& 69.9±3.5           & \underline {80.3±2.3}     & 61.7±2.7          & 64.5±1.9           & 71.6±1.9          & 67.0±3.0         							\\
NN                   & 88.9±2.2           & 87.5±2.5           & 51.9±7.0           & 69.1±2.6           & 67.6±3.3           & 52.1±6.3          							& 69.7±2.1           & 79.4±1.9           & 60.9±2.8          & 64.8±4.8           & 71.0±3.7          & 47.7±3.7          								\\
GRridge              & 80.6±2.7           & 78.2±3.3           & 88.3±2.8           & 62.7±2.8           & 61.9±4.0           & 68.6±3.4          							& 58.1±2.0           & 67.7±2.7           & 57.7±2.7          & 60.4±3.4           & 65.7±2.9          & 62.7±2.9          								\\
BSPLSDA              & 83.0±1.6           & 66.4±2.2           & 87.8±2.7           & 63.5±1.6           & 63.7±1.9           & 71.8±2.5          							& 58.3±1.7           & 67.7±2.3           & 60.2±2.8          & 61.4±2.7           & 69.3±3.0          & 62.5±2.6          								\\
GMU                  & 91.3±2.6           & 90.2±3.1           & 95.6±2.8           & 70.9±2.4           & 68.4±1.5           & 77.2±1.7          							& 68.4±2.7           & 78.3±2.8           & 58.6±2.7          & 68.5±2.3           & 64.8±3.3          & 74.0±2.2          								\\
Mogonet              & 95.3±1.3           & \underline {95.9±1.7}     & 91.4±1.5           & 78.6±2.1           & 77.0±1.7           & 82.7±2.2          					& 72.5±1.9           & 69.6±2.7           & \textbf{80.1±2.2} & 74.8±3.0           & 78.4±3.4          & \underline {78.1±3.5}    					\\
Dynamics             & \underline {96.5±0.6}     & 95.7±1.1           & 96.4±0.8           & \underline {80.4±1.6}     & \underline {78.9±1.6}     & \underline {84.1±1.8}  & \underline {75.8±1.6}     & 70.5±2.3           & \underline {78.3±1.7}    & \underline{76.9±2.6}     & \underline {79.7±2.7}    & 77.6±2.2           	\\
\textbf{Ours}        & \textbf{98.0±1.1*} & \textbf{97.7±1.3*} & \textbf{98.3±0.5*} & \textbf{83.7±1.2*} & \textbf{81.3±1.3*} & \textbf{85.9±1.6} 							& \textbf{78.6±1.5*} & \textbf{84.4±2.9*} & 77.3±1.2          & \textbf{80.0±2.0*} & \textbf{82.7±3.3} & \textbf{78.7±2.6} 						\\
\bottomrule
\addlinespace
\multicolumn{13}{p{480pt}} {Best is in \textbf{Bold}. Suboptimal results are \underline{underlined}. `\textasteriskcentered{}' indicates that our TMM is significantly better (t-test $p<0.05$) than the suboptimal method.}
\end{tabular}
\end{table*}

\begin{table*}[!thb]
\centering
\caption{Ablation study of the key components in TMM.}\label{tab:ablation_rri}
\begin{tabular}
{@{}>{\centering\arraybackslash}p{15pt}>{\centering\arraybackslash}p{20pt}|>{\centering\arraybackslash}p{25pt}>{\centering\arraybackslash}p{25pt}>{\centering\arraybackslash}p{25pt}|>{\centering\arraybackslash}p{25pt}>{\centering\arraybackslash}p{25pt}>{\centering\arraybackslash}p{25pt}|>{\centering\arraybackslash}p{25pt}>{\centering\arraybackslash}p{25pt}>{\centering\arraybackslash}p{25pt}|>{\centering\arraybackslash}p{25pt}>{\centering\arraybackslash}p{25pt}>{\centering\arraybackslash}p{25pt}}
\toprule
\multirow{3}{*}{T-RRI} & \multirow{3}{*}{R-RRI} & \multicolumn{3}{c|}{NC vs. AD} & \multicolumn{3}{c|}{NC vs. LMCI}& \multicolumn{3}{c|}{NC vs. EMCI} & \multicolumn{3}{c}{EMCI vs. LMCI} \\ 
\cmidrule{3-5}  \cmidrule{6-8}\cmidrule{9-11}  \cmidrule{12-14}
&& ACC & F1 & AUC & ACC & F1 & AUC & ACC & F1 & AUC & ACC & F1 & AUC  \\ \midrule 
\ding{51} & \ding{51} & \textbf{98.0±1.1} & \textbf{97.7±1.3} & \textbf{98.3±0.5} & \textbf{83.7±1.2} & \textbf{81.3±1.3} & \textbf{85.9±1.6} & \textbf{78.6±1.5} & \textbf{84.4±2.9} & \textbf{77.3±1.2} & \textbf{80.1±1.7} & \textbf{82.4±2.6} & \textbf{78.5±2.3}\\
\ding{51} & \ding{55} & 97.2±1.2 & 96.3±1.8 & 98.0±1.2 & 82.0±1.3 & 79.6±2.1 & 83.4±2.7 & 76.4±2.2 & 82.1±3.2 & 75.6±2.5 & 77.8±2.5 & 80.6±2.9 & 76.3±2.8\\
\ding{55} & \ding{51} & 96.7±1.4 & 95.9±2.1 & 97.2±1.4 & 81.4±1.6 & 78.2±1.4 & 83.6±1.8 & 77.3±1.4 & 82.8±2.7 & 76.8±1.6 & 78.3±2.3 & 81.8±2.4 & 77.5±2.0\\
\ding{55} & \ding{55} & 95.6±2.6 & 94.8±2.2 & 96.3±2.7 & 79.8±2.4 & 76.3±2.9 & 81.7±2.6 & 75.4±2.3 & 80.6±2.5 & 75.4±1.7 & 76.3±2.7 & 79.3±3.2 & 76.7±2.3\\
\bottomrule
\toprule
\multicolumn{2}{c|}{\multirow{3}{*}{Confidence}}  & \multicolumn{3}{c|}{NC vs. AD} & \multicolumn{3}{c|}{NC vs. LMCI}& \multicolumn{3}{c|}{NC vs. EMCI} & \multicolumn{3}{c}{EMCI vs. LMCI} \\ 
\cmidrule{3-5}  \cmidrule{6-8}\cmidrule{9-11}  \cmidrule{12-14}
&& ACC & F1 & AUC & ACC & F1 & AUC & ACC & F1 & AUC & ACC & F1 & AUC  \\ \midrule 
\multicolumn{2}{c|}{TFCP} & \textbf{98.0±1.1} & \textbf{97.7±1.3} & \textbf{98.3±0.5} & \textbf{83.7±1.2} & \textbf{81.3±1.3} & \textbf{85.9±1.6}  & \textbf{78.6±1.5} & \textbf{84.4±2.9} & \textbf{77.3±1.2} & \textbf{80.1±1.7} & \textbf{82.4±2.6} & \textbf{78.5±2.3} \\
\multicolumn{2}{c|}{TCP} & 97.2±1.1 & 97.3±1.1 & 98.1±0.6 & 81.4±1.6 & 80.2±1.4 & 83.2±2.0  & 78.3±1.2 & 83.6±2.6 & 77.0±0.8 & 79.5±1.4 & 82.1±1.9 & 78.2±2.2 \\
\multicolumn{2}{c|}{NN} & 96.6±1.6 & 96.0±2.0 & 97.3±1.6 & 79.5±2.3 & 78.5±2.6 & 81.5±2.7  & 77.4±1.5 & 82.2±2.7 & 76.3±1.4 & 77.4±2.4 & 80.1±2.8 & 75.2±2.6\\
\bottomrule
\end{tabular}
\end{table*}

\textbf{Benchmark methods.}
We compare TMM with nine multimodal competitors, including four single-modal classifiers with early fusion (SVM, Lasso, XGBoost, and fully connected NN), three models with intermediate concatenation (GRidge~\cite{van2016better}, BSPLSDA~\cite{singh2019diablo}, and GMU~\cite{arevalo2017gated}), and two methods with advanced representation and fusion designs (Mogonet~\cite{wang2021mogonet} and Dynamics~\cite{Han2022MultimodalDD}).

\textbf{Evaluation metrics.}
We employ accuracy (ACC), F1 score (F1), and the area under the receiver operating characteristic curve (AUC) to evaluate the performance of the methods. Five-fold cross-validation is performed to calculate the mean and standard deviation of the results. T-test is performed to evaluate the significance of improvements achieved by our method over state-of-the-art methods.

\textbf{Implementation details.}
We develop the model using PyTorch 2.1.0 and Adam as the optimizer. We train the model for 2,000 epochs with a learning rate of 1e-3 and weight decay of 1e-4. All experiments are implemented on an RTX 4090 GPU with 24GB of memory.

\subsection{Comparison with the state-of-the-art}
Four tasks, including NC vs. AD, NC vs. LMCI, NC vs. EMCI, and EMCI vs. LMCI, are designed to validate the performance of our proposed method. Since the transcriptomics knowledge is not specific to ADNI subjects and cannot be directly incorporated, we are limited to using only multimodal imaging data for the comparative methods. Table~\ref{tab:sota} shows the comparison results where our model illustrates superior performance across four tasks. Particularly, TMM achieves significant improvements (t-test $p<0.05$) in most metrics over the suboptimal results. This highlights the advantages of integrating molecular knowledge through the multi-view graphs and the proposed trustworthy strategy.

\subsection{Ablation studies}
We conduct extensive ablation studies to evaluate the effectiveness of including the transcriptomics-derived RRI knowledge, the disease-specific R-RRI networks and the TFCP module. We accordingly remove the T-RRI network and three R-RRI networks, and replace the TFCP with TCP and a fully connected NN. The ablation experiments are performed on four tasks. 
Table~\ref{tab:ablation_rri} shows the results, from which we can observe that: 1) Both T-RRI and R-RRI individually improve prediction performance, demonstrating that transcriptomics data and imaging data each offer unique insights into disease pathology; 2) Integrating transcriptomics with imaging data provides the best prediction outcomes, showcasing that biomedical data from varied perspectives can deliver effective complementary insights for AD; 3) Compared to TCP, the proposed TFCP shows enhanced performance, and similarly, TCP outperforms NN, suggesting that combining certainty and uncertainty yields more robust confidence estimates.

\begin{table*}[!htbp]
\footnotesize
  \centering
  \caption{Top 15 imaging biomarkers identified for each task.}
  \label{tab:biomarker}
  \begin{tabular}{@{}c@{\hskip 0.3cm}c@{\hskip 0.3cm}m{14cm}@{}}
\toprule
\textbf{Task} & \textbf{Modality} & \textbf{Top 15 ROIs$^a$}\\
\midrule 

\multirow{6}{*}{NC vs. AD} & VBM      & Amygdala\_L, Hippocampus\_R \\\addlinespace[3pt]
~    & FDG      & Cingulum\_Post\_L, ParaHippocampal\_L, ParaHippocampal\_R\\\addlinespace[3pt]
~     & AV45     & Hippocampus\_L, Hippocampus\_R, ParaHippocampal\_L, Cerebelum\_3\_L, Cerebelum\_3\_R, ParaHippocampal\_R, Cerebelum\_7b\_R, Cingulum\_Ant\_R, Temporal\_Pole\_Sup\_L, Cerebelum\_10\_L \\
\midrule
\multirow{6}{*}{NC vs. LMCI}   & VBM      & Hippocampus\_L, Amygdala\_L, Hippocampus\_R, Temporal\_Mid\_R\\\addlinespace[3pt]
~  & FDG      & Cingulum\_Post\_L, Cingulum\_Post\_R, ParaHippocampal\_L, Angular\_L, Angular\_R, Thalamus\_L, Occipital\_Inf\_L, Frontal\_Med\_Orb\_L \\\addlinespace[3pt]
~   & AV45     & ParaHippocampal\_L, Hippocampus\_R, Postcentral\_L \\
\midrule
\multirow{6}{*}{NC vs. EMCI}  & VBM      & Cerebelum\_10\_L, Hippocampus\_L, Hippocampus\_R, Amygdala\_L, Cerebelum\_9\_L  \\\addlinespace[3pt]              
~  & FDG      & Cingulum\_Post\_L, ParaHippocampal\_L, ParaHippocampal\_R, Hippocampus\_L, Angular\_L, Angular\_R, Cingulum\_Post\_R, Amygdala\_L, Occipital\_Mid\_L  \\\addlinespace[3pt]
~   & AV45     & Cerebelum\_3\_L  \\
\midrule
\multirow{4}{*}{EMCI vs. LMCI} & VBM      & Thalamus\_R, Supp\_Motor\_Area\_L, Hippocampus\_L, Hippocampus\_R \\\addlinespace[3pt]
~ & FDG      & Cingulum\_Post\_L, Putamen\_R, Putamen\_L, Cingulum\_Post\_R, Pallidum\_L, Vermis\_7, Pallidum\_R, Temporal\_Pole\_Mid\_R, Cerebelum\_8\_L, Vermis\_8, Vermis\_9\\
\bottomrule   
\addlinespace
\multicolumn{3}{p{460pt}} {$^a$The biomarkers are ranked by combining all three modalities, and the top 15 features for each classification task are listed.} 
  \end{tabular}
\end{table*}

\subsection{Evaluations of modality contributions} 
\begin{figure}[h]
    \centering
    \subfigure[]{\includegraphics[width=0.23\textwidth]{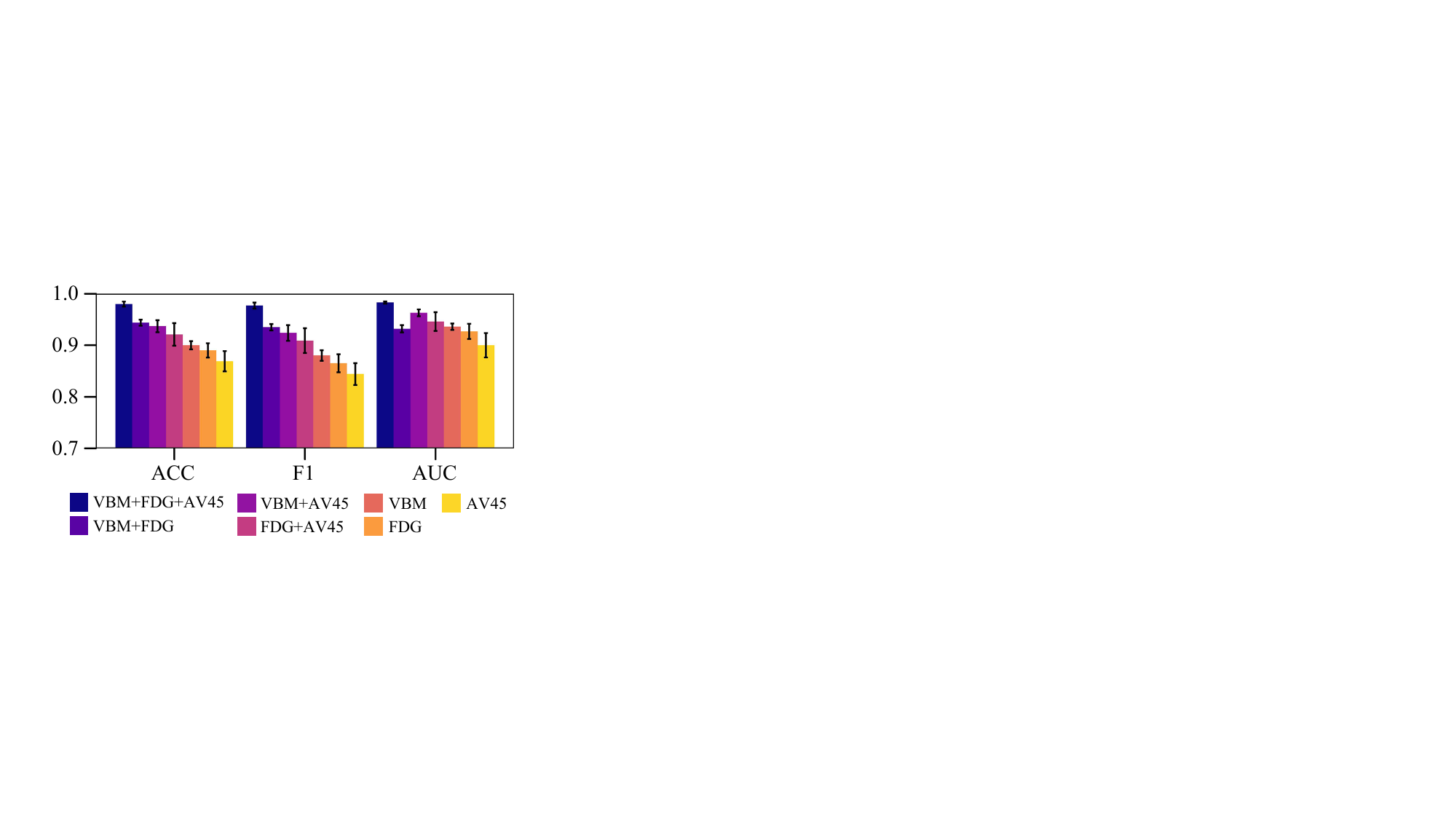}} 
    \subfigure[]{\includegraphics[width=0.21\textwidth]{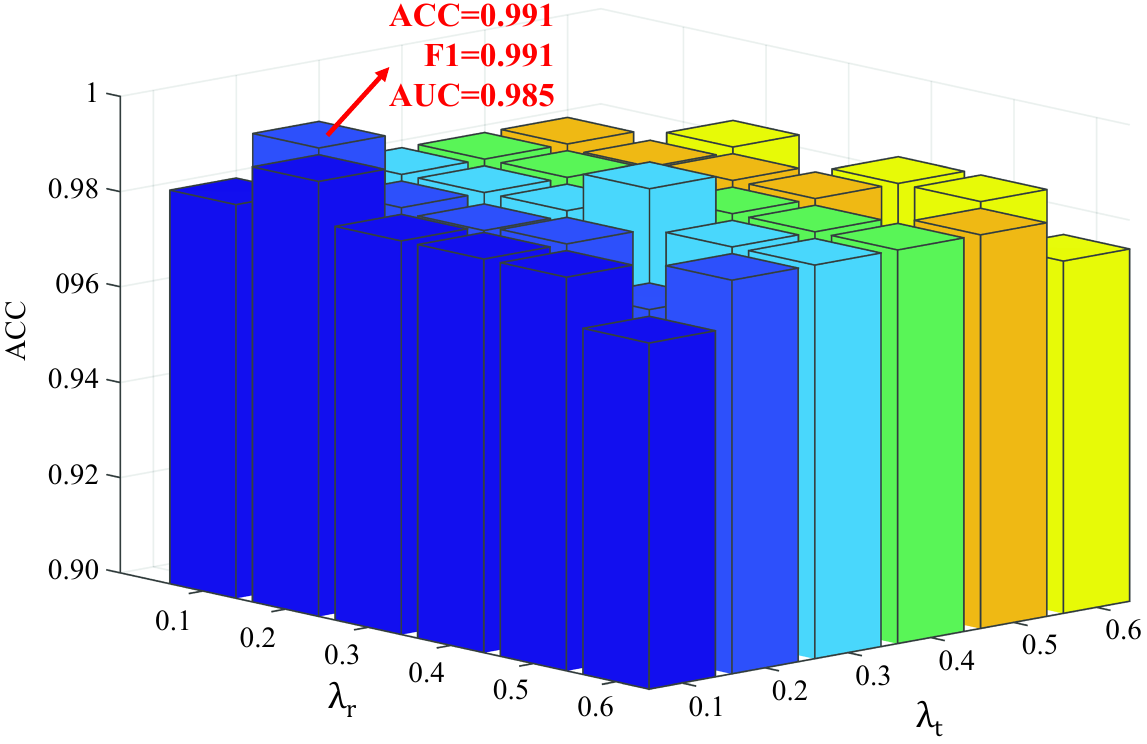}} 
    \caption{Performance comparison of (a) different modality combinations and (b) different hyperparameters for RRI thresholding on NC vs. AD task.}
    \label{fig:modal123_parameter}
\end{figure}

To illustrate how multiple imaging modalities can offer complementary insights, we compared the classification performance using various combinations of imaging data sources on the NC vs. AD task. The results, displayed in Figure~\ref{fig:modal123_parameter}(a), reveal that: 1) different imaging modalities possess varying levels of discriminative capabilities, with VBM outperforming both FDG and AV45; 2) the integration of additional imaging modalities consistently yields better performance than using any subset alone, highlighting the unique contribution of each imaging modality and affirming the ability of our model in effectively modeling cross-modal information.

\begin{figure}[!htbp]
    \centering
    \includegraphics[width=0.45\textwidth]{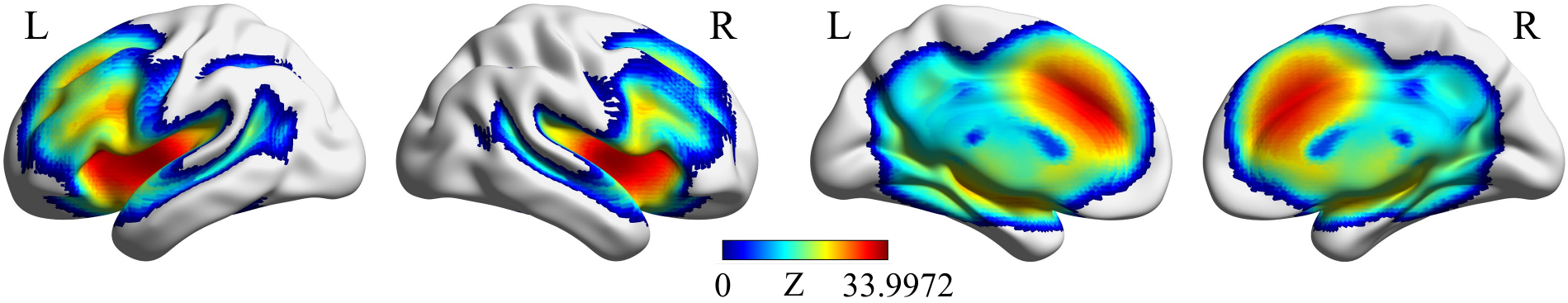}
    \caption{Coactivation maps of top ROIs obtained from NC vs. AD task. }
    \label{fig:coactivation}
\end{figure}

\subsection{Hyperparameter analysis.} 
We evaluate the hyperparameters for thresholding the T-RRI and R-RRI networks. The parameters $\lambda_t$ and $\lambda_r$ are automatically tuned from the NC vs. AD classification task within the range of $[0.1, 0.2, \cdots, 0.6]$. Figure~\ref{fig:modal123_parameter}(b) shows the grid search results, from which the optimal combination of $\lambda_t=0.2$ and $\lambda_r=0.1$ are selected in our experiments. Variations in the thresholding parameters have minimal impact on performance, demonstrating the robustness of RRI networks.

\begin{figure*}[!htbp]
    \centering
    \includegraphics[width=0.9\textwidth]{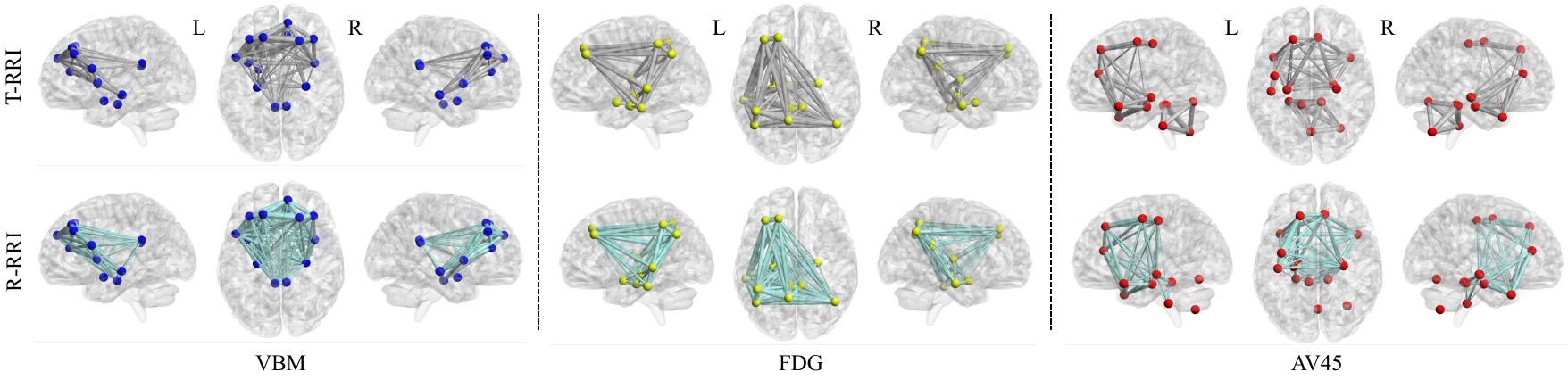}
    \caption{Transcriptomics-based (top panel, grey edges) and radiomics-based (bottom panel, green edges) connectivity of top ROIs derived from each modality. }
    \label{fig:rri}
\end{figure*}

\subsection{Identified biomarkers and functional analysis}
Feature ablation experiments~\cite{Dai_2021_WACV} are conducted to assess the importance of each imaging feature. Table~\ref{tab:biomarker} lists the top 15 biomarkers identified for each task. The hippocampus ranks highly in both early and late prediction tasks, aligning well with its recognition as one of the first regions to be affected by AD~\cite{deture2019neuropathological}. 
Among the top findings, several ROIs (e.g., left and right hippocampus, left posterior cingulum) are consistently present across various tasks, indicating their broad relevance. Some ROIs appear in most tasks (e.g., left amygdala, left parahippocampal gyrus), while some are specific to certain tasks (e.g., left thalamus in LMCI, right thalamus in EMCI vs. LMCI). This variability suggests that different regions may be involved at different AD stages.

We employ Neurosynth (\url{www.neurosynth.org}) to perform meta-analytic coactivation analysis, associating the identified ROIs with cognitive functions derived from 3,489 published neuroimaging studies. The coactivation map of top ROIs, as reported from NC vs. AD (refer to Table~\ref{tab:biomarker}), is illustrated in Figure~\ref{fig:coactivation}. The co-activated regions are primarily located in the subcortex (e.g., cingulate gyrus) and prefrontal lobe, which are known to be implicated in AD~\cite{pai2021time}.

We further visualize the connectivity of top ROIs underlying transcriptomics and each imaging modality. Specifically, we select the top 15 ROIs from each modality, extract their pairwise PCCs, and use the BrainNet Viewer (\url{https://www.nitrc.org/projects/bnv/}) for visualization. 
For each modality, distinct interaction patterns are observed between the transcriptomics-derived and radiomics-derived networks, illustrating how different biological levels offer unique insights from varying perspectives. This is especially marked in AV45, corresponding with the significant role of AV45 top ROIs (in NC vs. AD, 10 out of 15 ROIs are from the AV45) in classification. Further investigation is warranted into the identified ROIs and their role in AD progression.

\section{Conclusion}
In this paper, we propose TMM, a trustworthy enhanced multi-view multi-modal framework for AD prediction. We integrate upstream molecular knowledge with downstream radiomics information to effectively model relationships between brain ROIs and enhance sample representations. 
The proposed TFCP criterion successfully perceives sample-wise modal informativeness, facilitating dynamic fusion across multiple modalities.
Our experimental results across various AD classification tasks validate the efficacy of the proposed TMM. The identified biomarkers align with existing research and demonstrate relevance to AD progression.

\section*{Acknowledgment}

This work is partly supported by the National Natural Science Foundation of China (62102115, 62103116), the Fundamental Research Funds for the Central Universities (3072024GH2604), the Natural Science Foundation of Heilongjiang Province (LH2022F016), the Shandong Provincial Natural Science Foundation (ZR2024QF081, 2022HWYQ-093).

\bibliographystyle{IEEEtran}
\bibliography{reference}

\end{document}